\documentclass[conference]{IEEEtran}
\IEEEoverridecommandlockouts

\usepackage{cite}
\usepackage{hyperref}
\usepackage{amsmath,amssymb,amsfonts}
\usepackage{algorithmic}
\usepackage{graphicx}
\usepackage{textcomp}
\usepackage{xcolor}
\usepackage{bm}
\usepackage{multirow}
\usepackage{makecell}
\usepackage{colortbl}
\definecolor{c1}{HTML}{eecec1}
\definecolor{c2}{HTML}{740c59}
\definecolor{c3}{HTML}{89a8ae}

\usepackage{marvosym}  

\def\BibTeX{{\rm B\kern-.05em{\sc i\kern-.025em b}\kern-.08em
    T\kern-.1667em\lower.7ex\hbox{E}\kern-.125emX}}
\begin{document}

\title{Fresh-CL: Feature Realignment through Experts on Hypersphere in Continual Learning\\

\thanks{
This work is supported by the Sci-Tech Innovation Initiative by the Science and Technology Commission of Shanghai Municipality (24ZR1419000),  and the National Science Foundation of China (12471501).
}
}

\author{
\IEEEauthorblockN{\textit{Zhongyi Zhou$^1$, Yaxin Peng$^2$, Pin Yi$^2$, Minjie Zhu$^1$, Chaomin Shen $^1$\Letter}}
\IEEEauthorblockA{\\
$^1$School of Computer Science, East China Normal University, Shanghai, China\\
$^2$Department of Mathematics, Shanghai University, Shanghai, China}
}

\maketitle



\begin{abstract}
Continual Learning enables models to learn and adapt to new tasks while retaining prior knowledge. 
Introducing new tasks, however, can naturally lead to feature entanglement across tasks, limiting the model's capability to distinguish between new domain data.
In this work, we propose a method called Feature Realignment through Experts on hyperSpHere in Continual Learning (Fresh-CL). 
By leveraging predefined and fixed simplex equiangular tight frame (ETF) classifiers on a hypersphere, our model improves feature separation both intra and inter tasks.
However, the projection to a simplex ETF shifts with new tasks, disrupting structured feature representation of previous tasks and degrading performance. Therefore, we propose a dynamic extension of ETF through mixture of experts, enabling adaptive projections onto diverse subspaces to enhance feature representation.
Experiments on 11 datasets demonstrate a 2\% improvement in accuracy compared to the strongest baseline, particularly in fine-grained datasets, confirming the efficacy of combining ETF and MoE to improve feature distinction in continual learning scenarios.
\end{abstract}


\section{Introduction}

Humans are capable of achieving ongoing learning capacity and preservation of learned information. 
Similarly, deep learning models are expected to preserve past knowledge while continuously inferring new information. 
This domain, known as Continual Learning (CL), has gained significant attention in recent years. 
Existing works in CL continue to face a major challenge known as
catastrophic forgetting,
where the feature representations of previously learned data are overwritten by updates from newly introduced tasks. 
Despite numerous efforts, including changes in neural architecture (e.g., L2P \cite{b51}, DualPrompt \cite{b52}) and feature distillation (e.g., GFR \cite{b53}, FA \cite{b54} and 
DSR \cite{b55}), the problem in representation shift remains unresolved.
\begin{figure}
    \centering
    \includegraphics[width=1\linewidth]{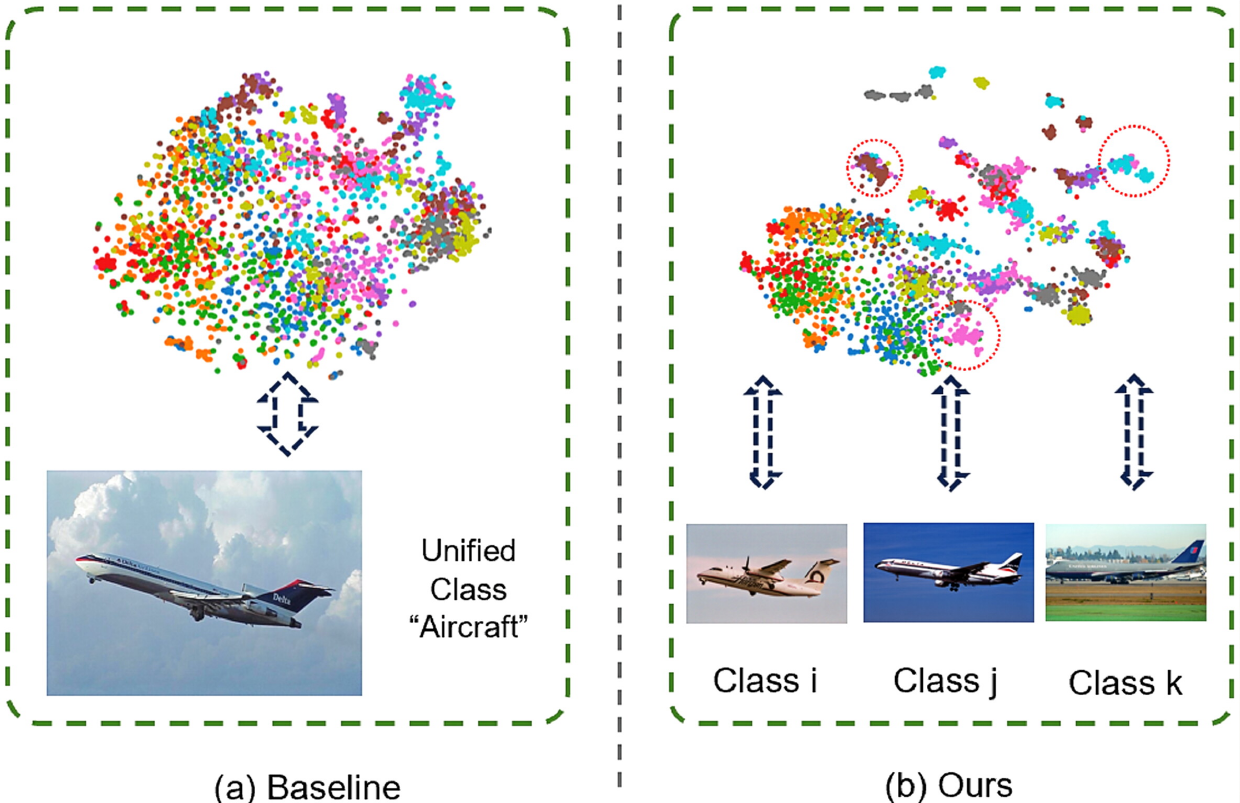}
    \caption{t-SNE visualization of Aircraft dataset features. 
    (a) Features extracted by MOA
    tend to blend together, making them recognizable as a single ``aircraft" class.
    (b) Our method improves feature separation, enabling distinct class recognition.}
    \label{tsne}
\end{figure}
One important observation that has been made for the catastrophic forgetting is that, deep learning models typically fail to disentangle feature representations across different classes, as the number of classes from diverse domains incrementally added into the model. This especially happens in fine-grained datasets and can lead to feature overlap and representation confusion, as is shown in Fig.~\ref{tsne}.

To address this challenge, recent works have turned to neural collapse (NC), a phenomenon in which classifier vectors and activations for all classes converge into a simplex equiangular tight frame (ETF)\cite{b1,b5,b8,b15}. In this structure, all classes are represented as maximally separated points, effectively reducing the overlap between class features. 
While NC improves feature separation in intra-domain tasks \cite{b44,b45,b47}, its performance diminishes in inter-domain tasks.
As new tasks with significant domain differences are introduced, the projection onto the ETF degrades, disrupting the structure learned from earlier tasks and causing performance drop across domains.

In this paper, we propose a neural collapse inspired mixture of experts model, \ termed as Feature Realignment through Experts on hyperSpHere in Continual Learning (Fresh-CL). By using predefined equiangular vectors as targets for each class feature, features across different classes can be distinguished effectively. However, we found that the model's ability to maintain projection onto the ETF may gradually diminish when only a single ETF is used. To address this, we propose a mixture-of-experts approach that leverages multiple ETFs to improve the classification of feature representations across different classes.

\begin{figure*}
    \centering
    \includegraphics[width=1\linewidth]{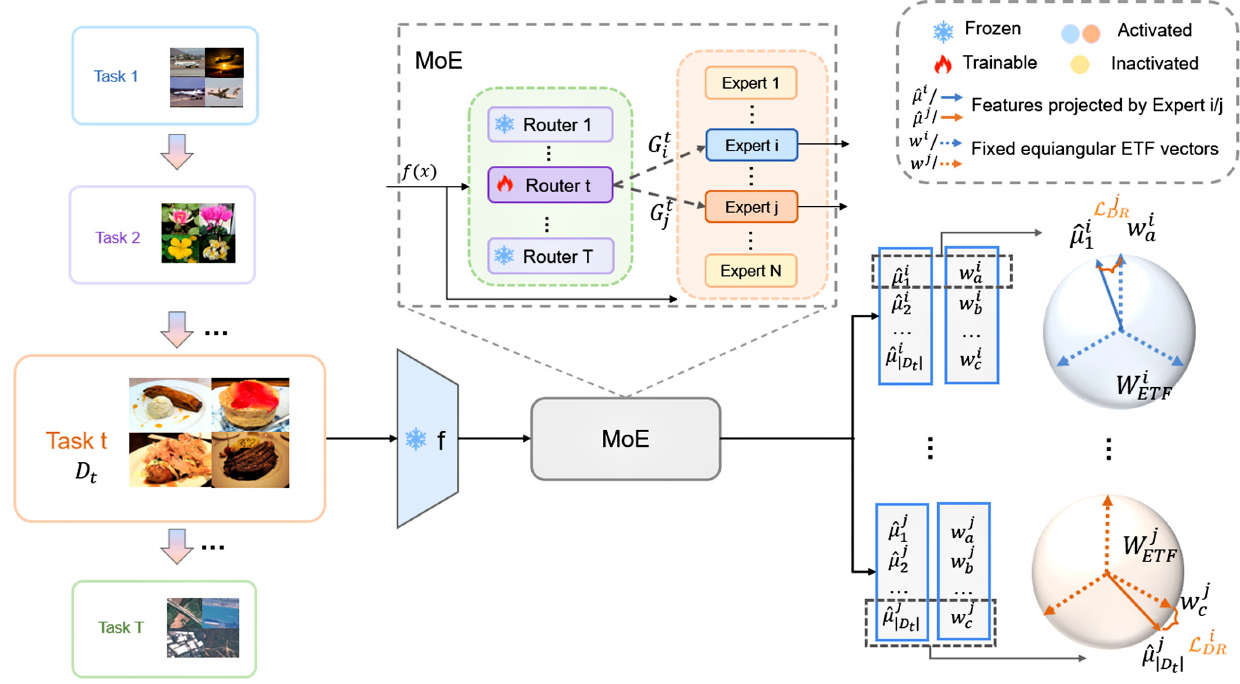}
    \caption{Overall framework of Fresh-CL. When training \textbf{Task $\bm t$}, each input image $\boldsymbol x_{k}$ is fed into a frozen backbone, and then processed through a dynamic routing mechanism, which select the top-$k$ (e.g. top-2) appropriate experts $i$ and $j$ using gating weights $G_i^t$ and $G_j^t$. These experts project the normalized features $\hat{\boldsymbol{\mu}}_{k}^{i}$ and $\hat{\boldsymbol{\mu}}_{k}^{j}$ into distinct hyperspheres, aligning each of them with their corresponding predefined ``pseudo targets" $\bm w_{a}^i$ and $\bm w_{a}^j$ on the relevant hypersphere. The subscripts $a, b, c$ represent arbitrary labels from the label space $C^t$. These pseudo targets are predefined as equiangular vectors $\hat{\bm{W}}_{ETF}^{i}$ and $\hat{\bm{W}}_{ETF}^{j}$ for each class of all tasks. By using the DR loss, we force these features close to their corresponding targets, therefore achieve max separation on each appropriate hypersphere selected by the MoE module.}
    \label{framework}
\end{figure*}

Our contribution can be summarized as follows:
\begin{itemize}
\item We introduce an optimal pre-assigned alignment structure, inspired by neural collapse, as fixed targets, aiming to distinguish feature representations from each other in multi-domain tasks.
\item We propose an extensible framework which employs experts to provide multiple projections to different hyperspheres, inducing neural collapse in various subspaces.
\item Experiments on 11 datasets illustrate that our method achieves 2\% and 1.1\% over a recent strong baseline under full-shot and few-shot multi-task incremental learning settings.
\end{itemize}

\section{Preliminaries}
\subsection{Problem Statement of Continual Learning}
Existing continual learning (CL) methods are typically categorized into Class Incremental Learning (CIL) 
\cite{b12,b16,b18}, which learns new classes within a single domain, and Task Incremental Learning (TIL)
\cite{b20,b21,b22}, which addresses tasks across different domains. Specifically, each CL method trains a model incrementally on a sequence of training datasets 
$\left\{\mathcal{D}^{(0)},\mathcal{D}^{(1)},\cdots,\mathcal{D}^{(T)}\right\}$, where 
$\mathcal{D}^{(t)}=\{(\boldsymbol{x}_{i},y_{i})\}_{i=1}^{|\mathcal{D}^{(t)}|}$ 
for task $t \in [0,T]$, with $\boldsymbol{x}_i$ representing the input image and $y_i$ representing the corresponding label. 
The label space $\mathcal{C}^{(t)}$ of task $t$ has no overlap with any other task $t^{\prime}$, i.e. $\mathcal{C}^{(t)}\cap\mathcal{C}^{(t^{\prime})}=\emptyset$, for all $ t^{\prime}\neq t\in [0,T]$. When learning task $t$, data of previous tasks remains inaccessible. Only the current dataset $\mathcal{D}^{(t)}$ can be accessed.

\subsection{Neural Collapse}
Previous studies \cite{b8} observed that the last layer feature and its corresponding classifier weights form an optimal structure at the terminal phase of training. This phenomenon is called neural collapse.

When neural collapse happens, the collection of $K$ classifier vectors $\bm{w}_1,\bm{w}_2,\cdots,\bm w_K$ form a simplex equiangular tight frame (ETF) $\bm{W}_{ETF}=[\bm{w}_1,\bm{w}_2,\cdots,\bm w_K]$, which satisfies 
\begin{align}
    \bm{w}_{i}^T\bm{w}_{j}=\frac{K}{K-1}\delta_{i,j}-\frac{1}{K-1}, \quad \forall i,j\in[1,K].
\end{align}
where $\delta_{i,j}=1$ when $i=j$, and 0 otherwise.

This implies that all column vectors in $\bm{W}_{ETF}$ have the same $l_2$ norm and each pair has the same inner product.
\begin{table*}[htbp]
\caption{Last accuracy score under full-shot MTIL setting.}
\vspace{-20pt}
\begin{center}
\resizebox{\textwidth}{!}{
\footnotesize
\begin{tabular}{c|c|c|c|c|c|c|c|c|c|c|c|c}
\hline
\textbf{Method} 
&  \textcolor{c2}{\textbf{\textit{\rotatebox{25}{\makecell{Aircraft$\star$ \\ \cite{b25}}}}}}
& \textbf{\textit{\rotatebox{25}{\makecell{Caltech101\\ \cite{b30}}}}}
& \textbf{\textit{\rotatebox{25}{\makecell{CIFAR100\\ 
\cite{b31}}}}} 
& \textbf{\textit{\rotatebox{25}{\makecell{DTD\\ \cite{b32}}}}}
& \textbf{\textit{\rotatebox{25}{\makecell{EuroSAT\\ \cite{b33}}}}}
& \textcolor{c2}{\textbf{\textit{\rotatebox{25}{\makecell{Flowers$\star$\\ \cite{b26}}}}}}
& \textcolor{c2}{\textbf{\textit{\rotatebox{25}{\makecell{Food$\star$\\ \cite{b27}}}}} }
& \textbf{\textit{\rotatebox{25}{\makecell{MNIST\\ \cite{b34}}}}}
& \textcolor{c2}{\textbf{\textit{\rotatebox{25}{\makecell{OxfordPet$\star$\\ \cite{b28}}}}}} 
& \textcolor{c2}{\textbf{\textit{\rotatebox{25}{\makecell{StanfordCars$\star$\\ \cite{b29}}}}}}
& \textbf{\textit{\rotatebox{25}{\makecell{SUN397\\ \cite{b35}}}}} & \textbf{Average}
\\
\hline

Continual-FT    &31.0   &89.3   &65.8   &67.3   &88.9   &71.1   &85.6   &\textbf{99.6}&92.9&77.3&81.1&77.3 \\
LwF            &26.3   &87.5   &71.9   &66.6   &79.9   &66.9   &83.8   &\textbf{99.6}&92.1&66.1&80.4&74.6\\
iCaRL          &35.8   &93.0   &77.0   &70.2   &83.3   &88.5   &90.4   &86.7       &93.2&81.2&\textbf{81.9}&80.1\\
LwF-VR          &20.5   &89.8   &72.3   &67.6   &85.5   &73.8   &85.7   &\textbf{99.6}&93.1&73.3&80.9&76.6\\
WiSE-FT        &27.2   &90.8   &68.0   &68.9   &86.9   &74.0   &87.6   &\textbf{99.6}&92.6&77.8&81.3&77.7\\
ZSCL           &40.6   &92.2   &81.3   &70.5   &94.8   &90.5   &91.9   &98.7   &93.9&85.3&80.2&83.6\\
MOA            &49.8   &92.2   &\textbf{86.1}  &78.1   &95.7   &94.3   &89.5   &98.1&89.9&81.6&80.0&85.0\\

\rowcolor{c1!70}
\textbf{Fresh-CL(1k)}&51.6&94.9&83.6&77.9&95.3&97.0&92.1&98.8&\textbf{94.1}&86.2&80.3&86.5(+1.5)\\
\rowcolor{c1!70}
\textbf{Fresh-CL(3k)}
&\textbf{52.4}  &\textbf{95.4}  &83.5   &\textbf{78.2}  &\textbf{96.9}  &\textbf{97.1}  &\textbf{92.7}  &98.8&93.8  &\textbf{86.5}  &81.2   &\textbf{87.0(+2.0)}\\
\hline
\multicolumn{13}{l}{
$\star$ represent fine-grained datasets.}
\end{tabular}}
\label{tab1}
\end{center}
\end{table*}
\section{Methodology}

In this paper, we propose Fresh-CL, an extensible training framework designed to enhance feature separation across various continual learning settings. The overall architecture of our proposed framework is illustrated in 
Fig.\,\ref{framework}. In Section \ref{3.1}, we first present our fixed simplex equiangular tight frame (ETF) classifier, which is inspired by the neural collapse phenomenon. Beyond the standard projection which maps onto the same hypersphere, we introduce a series of specialized experts. These experts are coupled with a gating mechanism that dynamically adjusts the projection pathways. This design ensures task-specific alignment between features and their respective hypersphere targets, as elaborated in Section~\ref{3.2}.

\subsection{Inducing Neural Collapse for Continual Learning}\label{3.1}

While the common models incrementally introduce new tasks or classes, class separation may gradually decreases and feature overlap may occur, leading to performance degradation.

Considering about the optimal alignment provided by ETF structure, which ensures that classes are maximally separated in the feature space, we are motivated to induce neural collapse in a CL process.

Specifically, we employ a fixed ETF $\hat{\bm{W}}_{ETF} \in \mathbb{R}^{d \times K}$ as a ``pseudo target" on the hypersphere, where $d$ is the dimension of the feature space and $K$ is the number of ETF vectors. 
The pre-defined pseudo targets are generated by the ETF structure. Although these targets do not directly correspond to specific class features, they serve as idealized reference points.
These targets maintain maximum separation on the hypersphere. As a result, they provide a stable reference point, which helps prevent feature interference when learning new classes.
This is particularly important in CL, where newly introduced tasks may interfere with previously learned class representations.
 
In particular, when incrementally training new data from novel classes or tasks, the extracted feature is guided by Dot-Regression Loss ($\mathcal{L}_{DR}$). This loss function aligns the feature with the pseudo target vector on the hypersphere, which is indexed by the class-encoded numerical value, i.e.
\begin{align}
    \mathcal{L}_{DR}\left(\hat{\boldsymbol{\mu}}_{i},\hat{\bm{W}}_{ETF}\right)=\frac{1}{2}\left(\hat{\bm{w}}_{{y_{i}}}^{T}\hat{\boldsymbol{\mu}}_{i}-1\right)^{2},
\end{align}
where $y_{i}\in \mathbb{R^+}$ is the class label of data $\bm x_{i}$, 
$\hat{\bm{w}}_{y_i}$ is the $y_i$-th vector of $\hat{\bm{W}}_{ETF}$,
and
$\hat{\boldsymbol{\mu}}_{i}=\boldsymbol{\mu}_{i}/\|\boldsymbol{\mu}_{i}\|$ is the normalized feature of $\bm x_{i}$ extracted by the backbone $f$.

A key consideration is the dimension of $\hat{\bm{W}}_{ETF} \in \mathbb{R}^{d \times K}$. 
Unlike previous works that pre-counted the total number of classes for all tasks to specify a fixed $K$, we do not assume a known total class number. Instead, we set $K$ equal to the feature dimension $d$, since the total number of classes is unknown and will increase in real CL scenarios.

\subsection{Incremental Mixture of ETF Classifiers}\label{3.2}

Although a single ETF can effectively distinguish features across different classes, when the differences between task domains are significant, the way the model projects onto the ETF may gradually be forgotten. In other words, while the new domain can be accommodated through adaptive projection, this often comes at the cost of disrupting the ETF properties learned from earlier tasks. As a result, the model’s performance on previous domains may degrade. 
To achieve the goal of domain-specific adaption, we introduce Mixture of Experts (MoE) \cite{b23,b24} into our framework. By dynamically selecting experts tailored to each task, MoE enables different projections for tasks across different domains, ensuring that each task is mapped to the appropriate subspace.

\textbf{ETF classifier as Expert.}
MoE is designed to improve model performance by dynamically selecting specialized sub-networks, named ``experts", based on the input data. Each expert is a fully functional neural network, trained to perform well on specific subsets of the input space. 
As previously mentioned, a predefined ETF is employed as a pseudo target, enabling the projection layer to realign features with this target, thereby promoting effective feature separation.

Building upon this approach, each expert in our framework is designed as a specialized projection layer associated with a dedicated ETF. The ETF classifier expert ensures the maximal class separation by aligning features with predefined equiangular targets on the hypersphere, thus effectively addressing task-specific feature representation needs.
This design ensures that each task is mapped onto a distinct hyperspherical subspace, allowing for task-specific feature representations.

\begin{table*}[htbp]
\caption{Last accuracy score under few-shot MTIL setting}
\vspace{-20pt}
\begin{center}
\resizebox{\textwidth}{!}{
\footnotesize
\begin{tabular}{c|c|c|c|c|c|c|c|c|c|c|c|c}
\hline
\textbf{Method} 
&  \textcolor{c2}{\textbf{\textit{\rotatebox{25}{\makecell{Aircraft$\star$ \\ \cite{b25}}}}}}
& \textbf{\textit{\rotatebox{25}{\makecell{Caltech101\\ \cite{b30}}}}}
& \textbf{\textit{\rotatebox{25}{\makecell{CIFAR100\\ 
\cite{b31}}}}} 
& \textbf{\textit{\rotatebox{25}{\makecell{DTD\\ \cite{b32}}}}}
& \textbf{\textit{\rotatebox{25}{\makecell{EuroSAT\\ \cite{b33}}}}}
& \textcolor{c2}{\textbf{\textit{\rotatebox{25}{\makecell{Flowers$\star$\\ \cite{b26}}}}}}
& \textcolor{c2}{\textbf{\textit{\rotatebox{25}{\makecell{Food$\star$\\ \cite{b27}}}}} }
& \textbf{\textit{\rotatebox{25}{\makecell{MNIST\\ \cite{b34}}}}}
& \textcolor{c2}{\textbf{\textit{\rotatebox{25}{\makecell{OxfordPet$\star$\\ \cite{b28}}}}}} 
& \textcolor{c2}{\textbf{\textit{\rotatebox{25}{\makecell{StanfordCars$\star$\\ \cite{b29}}}}}}
& \textbf{\textit{\rotatebox{25}{\makecell{SUN397\\ \cite{b35}}}}} & \textbf{Average}
\\
\hline
Continual-FT&27.8&86.9&60.1&58.4&56.6&75.7&73.8&93.1&82.5&57.0&66.8&67.1\\
LwF&22.1&58.2&17.9&32.1&28.1&66.7&46.0&84.3&64.1&31.5&60.1&46.5\\
LwF-VR&22.9&89.8&59.3&57.1&57.6&79.2&78.3&77.7&83.6&60.1&69.8&66.9\\
WiSE-FT&30.8&88.9&59.6&60.3&80.9&81.7&77.1&\textbf{94.9}&83.2&62.8&70.0&71.9\\
ZSCL&26.8&88.5&63.7&55.7&60.2&82.1&82.6&58.6&85.9&66.7&70.4&67.4\\
MOA&30.1&89.0&68.9&\textbf{63.7}&\textbf{82.2}&88.7&84.9&89.1&87.8&69.6&72.3&75.1\\
\rowcolor{c1!70}
\textbf{Fresh-CL}&\textbf{31.8}&\textbf{90.4}&\textbf{69.5}&62.5&81.7&\textbf{94.2}&\textbf{86.8}&88.6&\textbf{89.2}&\textbf{70.1}&\textbf{72.9}&\textbf{76.2(+1.1)}\\
\hline
\multicolumn{13}{l}{$\star$ represent fine-grained datasets.}
\end{tabular}
\label{tab2}}
\end{center}
\end{table*}

\textbf{Incremental Router for Expert Selection.}
Regular setting in MoE uses one router to decide the probability of using certain expert. Under continual learning setting, we introduce a task-specific router $\bm{R}^{t}$ for each task $t$, which dynamically computes the probability $\bm G^{t}$ for selecting the most appropriate experts. The router ensures that each task is assigned to the experts best equipped to address its distinct requirements. Therefore, for each data $\bm x_k$, the overall loss function can be written as
\begin{align}
    \mathcal{L}_{k}^t=\sum_{i=1}^{N_E}G_i^t\mathcal{L}_{DR}\left(\hat{\boldsymbol{\mu}}_{i}, \bm W_{ETF}^{i}\right),
\end{align}
where 
$G_i^t$ is the $i$-th column of $\bm G^t$, representing the gating weights of experts and can be computed as
\begin{align}
    \bm G^t=\operatorname{Softmax}(\operatorname{Topk}(\bm R^{t}(\hat{\boldsymbol{\mu}}_{i}))).
\end{align}

The $\operatorname{Topk}()$ method selects the $k$ most relevant experts, and $\operatorname{Softmax}()$ normalizes the resulting values into a probability distribution over the selected experts. The gating network is trained to assign tasks to the appropriate experts, ensuring that task-relevant features are enhanced while reducing interference from other tasks.

After completing each training task $t$, we freeze the top $k$ most frequently utilized experts to retain their domain-specific knowledge and specialized projection methods. These frozen experts provide stable feature representations for subsequent tasks, enhancing robustness. Despite being frozen, these experts remain selectable in future tasks, allowing the model to leverage their learned representations without additional updates.

\section{Experiments}
\subsection{Experimental Setting}
We evaluate Fresh-CL on 11 datasets under the Multi-task Incremental Learning (MTIL)\cite{b22} setting. These datasets include fine-grained datasets (e.g. Aircraft\cite{b25}, Flowers\cite{b26}, Food\cite{b27}, OxfordPet\cite{b28}, StanfordCars\cite{b29}), and coarse-grained datasets (e.g. Caltech101\cite{b30}, CIFAR100\cite{b31}, DTD\cite{b32}, EuroSAT\cite{b33}, MNIST\cite{b34}, SUN397\cite{b35}).

To evaluate our method in mitigating the forgetting of previous knowledge, we use the last accuracy ($A_{last}$) to measure the accuracy at the end of training across all tasks. We follow the training order proposed in \cite{b22}.

\textbf{Implementation Details}
We use the CLIP encoder with ViT-B/16\cite{b36} as the backbone to extract image feature and set the total number of experts $N_{E}=22$. The router is a single MLP that activates the experts with the top-2 gating scores. Following \cite{b37}, we train an auto-task identifier using the
pretrained AlexNet incrementally, therefore obtain the pseudo task id when performing inference on id-agnostic tasks. We train our method for 1k iterations per task under both full-shot and few-shot settings, using AdamW as our optimizer.

\subsection{Comparison with State-of-the-art Methods}
We compare our Fresh-CL with various CL methods, including LwF\cite{b38}, iCaRL\cite{b42}, LwF-VR\cite{b39}, WiSE-FT\cite{b40}, ZSCL\cite{b41} and MOA\cite{b37}.

\textbf{Full shot MTIL.}
Tab. \ref{tab1} displays the performance between Fresh-CL and other methods under the full shot MTIL setting. All the datasets in Tab.\,\ref{tab1} are trained and tested following the left to right order. Fresh-CL outperforms the challenging baseline MOA by 1.5\% in average, especially performs well in fine-grained datasets. By increasing the training iterations to 3k, our method (labeled ``Fresh-CL(3k)") achieves further improvement of 2\%.

\textbf{Few shot MTIL.}
Tab. \ref{tab2} shows our results under the few-shot MTIL setting, where the model is trained with 5 samples per class. Our Fresh-CL outperforms the second-best approach by an average of 1.1\%, demonstrating its robustness in this challenging scenario.


\textbf{Computation Cost.}
Thanks to the sparse property of MoE, our model only has trainable parameters of 56.22MB. 
\vspace{-0.3cm}

\setlength{\belowcaptionskip}{100pt}  
\begin{table}[!htbp]
\caption{Ablation Study on incremental ETF classifiers}
\vspace{-20pt}
\footnotesize
\begin{center}
\resizebox{0.5\textwidth}{!}{
\begin{tabular}{c|c|c|c}
\hline
\multirow{2}*{\textbf{Method}}&\multicolumn{2}{c|}{\textbf{Parameters}}& \multirow{2}*{\textbf{\textit{Average}}}
\\
\cline{2-3}
 ~& \textbf{Expert Number} & \textbf{Router Number} & ~  \\
\hline
Continual-FT & / & / &77.3\\
Fixed ETF w/o MoE & / & / & 53.5\\
Fixed ETF & 2 & 1 & 58.24\\
Fixed ETF & 2 & 11 & 62.9\\
Fixed ETF & 4 & 11 & 71.3\\
Fixed ETF & 8 & 11 & 73.3\\
Fixed ETF & 11 & 11 & 84.4\\

\hline

\end{tabular}}
\label{tab3}
\end{center}
\end{table}
\vspace{-0.5cm}
\subsection{Ablation Study}
We conduct an ablation study as shown in Tab. \ref{tab3}. All of these methods are built upon the same framework. To ensure fairness in scenarios with a limited number of experts, a probability of number of experts/11 is applied to freeze experts after the training stage for each task. Tab. \ref{tab3} indicates that the degree of feature forgetting is progressively reduced as the number of experts increases, effectively mitigating the representation shift.

\section{Conclusion}
We have 
introduced Fresh-CL, an NC-inspired model to mitigate feature interference in continual learning. By utilizing fixed simplex ETF as pseudo targets and incorporating a dynamic MoE, our approach has enhanced feature separation across tasks, ensuring distinct representations and mitigating the forgetting problem. 
Extensive experiments has shown that Fresh-CL outperforms state-of-the-art methods in both full-shot and few-shot settings, demonstrating its effectiveness in continual learning scenarios. 


\vspace{12pt}

\end{document}